\documentclass{Interspeech}

\interspeechcameraready 

\usepackage{comment}
\usepackage{lipsum}
\usepackage{amsmath,amssymb,amsfonts}
\usepackage{algorithmic}
\usepackage{graphicx}
\usepackage{textcomp}
\usepackage{booktabs}
\usepackage{tabularx}
\usepackage{arydshln}
\usepackage{bm}
\usepackage{multirow,makecell}
\usepackage[dvipsnames]{xcolor}
\usepackage{soul}
\usepackage[
 bibencoding=utf8,
 backend=biber,
 bibstyle=ieee,
 citestyle=numeric-comp,
 defernumbers=true,
 sortcites=true,
 maxbibnames=20,
 minbibnames=4,
]{biblatex}
\usepackage[export]{adjustbox}
\usepackage[caption=false,font=footnotesize]{subfig}
\usepackage{hyperref}
\hypersetup{
    colorlinks=true,
    linkcolor=blue,
    urlcolor=blue,
    citecolor=blue,
}
\usepackage[nameinlink, capitalise]{cleveref}
\usepackage[dvipsnames]{xcolor}

\definecolor{light_blue}{RGB}{0, 90, 181}
\definecolor{light_red}{RGB}{220, 50, 32}

\usepackage{pifont}
\newcommand{\cmark}{\ding{51}}%
\newcommand{\xmark}{\ding{55}}%

\newcommand\blfootnote[1]{%
  \begingroup
  \renewcommand\thefootnote{}\footnote{#1}%
  \addtocounter{footnote}{-1}%
  \endgroup
}

\def\BibTeX{{\rm B\kern-.05em{\sc i\kern-.025em b}\kern-.08em
    T\kern-.1667em\lower.7ex\hbox{E}\kern-.125emX}}

\title{Task Arithmetic for Target Language Expansion in Speech Translation}

\author[affiliation={1,{\dagger}}]{Yao-Fei}{Cheng}
\author[affiliation={2}]{Hayato}{Futami}
\author[affiliation={2}]{Yosuke}{Kashiwagi}
\author[affiliation={2}]{Emiru}{Tsunoo}
\author[affiliation={3,{\dagger}}]{Wen Shen}{Teo}
\author[affiliation={4}]{\\Siddhant}{Arora}
\author[affiliation={4}]{Shinji}{Watanabe}

\affiliation{}{University of Washington}{USA}
\affiliation{}{Sony Group Corporation}{Japan}
\affiliation{}{University of Electro-Communications}{Japan}
\affiliation{}{Carnegie Mellon University}{USA}
\email{
    nlp5566@uw.edu
}


\begin{document}

\maketitle

\begin{abstract}

Recent progress in large language models (LLMs) have gained interest in speech-text multimodal foundation models, achieving strong performance on instruction-tuned speech translation (ST).
However, expanding language pairs is costly due to re-training on combined new and previous datasets.
To address this, we aim to build a one-to-many ST system from existing one-to-one ST systems using task arithmetic without re-training.
Direct application of task arithmetic in ST leads to language confusion; therefore, we introduce an augmented task arithmetic method incorporating a language control model to ensure correct target language generation.
Our experiments on MuST-C and CoVoST-2 show BLEU score improvements of up to 4.66 and 4.92, with COMET gains of 8.87 and 11.83.
In addition, we demonstrate our framework can extend to language pairs lacking paired ST training data or pre-trained ST models by synthesizing ST models based on existing machine translation (MT) and ST models via task analogies.
\end{abstract}

\section{Introduction}
\label{sec:intro}

Following the recent development of large language models (LLMs) \cite{gemini_gemini_2024, openai_gpt-4_2024, dubey_llama_2024}, large speech-text multimodal foundation models, such as SpeechGPT \cite{zhang_speechgpt_2023}, have gained rapid attention \cite{rubenstein_audiopalm_2023, fathullah_prompting_2023, maiti_voxtlm_2024}.
Similar to automatic speech recognition (ASR) \cite{du_lauragpt_2023} and spoken language understanding (SLU) \cite{gong_joint_2023,arora_universlu_2024} tasks, speech translation (ST) \cite{wu_decoder-only_2023, huang_investigating_2024} has seen significant performance improvements based on multimodal foundation models.
Usually, such ST models only support translation language pairs that have been seen in training.
In practical usage, there may be a need to translate previously unseen language pairs \cite{qian_learn_2024}.
Therefore, extending support for new language pairs beyond those available in an existing ST model is essential.
However, this usually involves re-training the model with a combination of new and existing ST datasets, increasing computational costs as the training data grows.
Alternatively, training an ST model exclusively on the new language pair and integrating it with an existing pre-trained ST model, without the need for re-training, can significantly reduce training costs.
To this end, we introduce a task arithmetic-based \cite{ilharco_editing_2023} model merging framework to achieve this integration.


\begin{figure}
    \centering
    \includegraphics[width=0.45\textwidth]{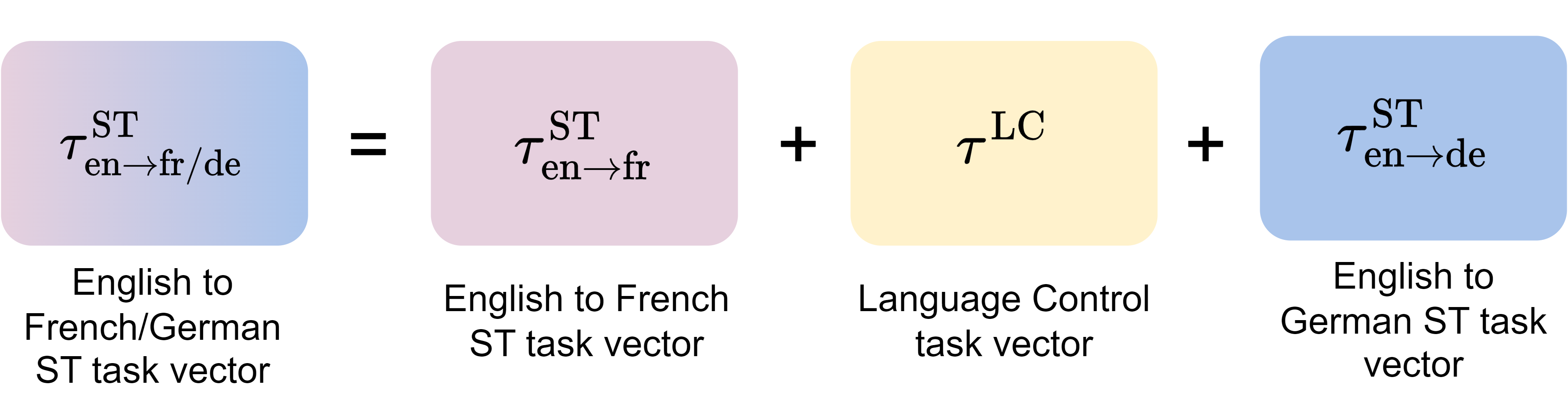}
    \caption{Task arithmetic (task vector addition) on target language expansion to achieve one-to-many ST models. Please refer to \cref{sec:method} for the detailed definition of notations.}
    \label{fig:merging}
    \vspace{-20pt}
\end{figure}







Model merging, a training-free approach, has been proposed to get a new model by directly manipulating the parameters of existing models \cite{wortsman_model_2022,matena_merging_2022,jin_dataless_2023,ilharco_editing_2023,gueta_knowledge_2023,goddard_arcees_2024,akiba_evolutionary_2024,biggs_diffusion_2024}.
The popular model merging approach, task arithmetic \cite{ilharco_editing_2023}, has shown that task expansion can be achieved by task vector addition.
A task vector is computed as the difference between the fine-tuned and pre-trained model parameters.
For further performance improvement, advanced methods built on top of task arithmetic have been proposed to solve parameter interference by pruning the parameters and aggregating task vectors during merging \cite{yadav_ties-merging_2023,yu_language_2024}.
Task arithmetic and its extensions have been applied to various tasks, such as image classification \cite{ilharco_editing_2023, yadav_ties-merging_2023} and generation \cite{biggs_diffusion_2024}, text classification \cite{yadav_ties-merging_2023}, text-to-speech (TTS) \cite{murata_attribute_2024}, and ASR \cite{ramesh_task_2024}.
While it has been applied in various domains, multilingual text generation in ST has not been explored.

This work explores a target language expansion aiming to build a one-to-many ST system from existing one-to-one ST systems using task arithmetic without re-training, as illustrated in \cref{fig:merging}.
As a target language expansion example, we build En $\rightarrow$ \{De, Fr\} or En $\rightarrow$ \{De, Zh\} systems by using existing En $\rightarrow$ Fr, En $\rightarrow$ De, and En $\rightarrow$ Zh systems.
We find directly applying task arithmetic leads to substantial language confusion errors, where the models generate translations in the wrong target language.
As a remedy, we propose to merge\footnote{We use the term merge to refer to performing task arithmetic in this work.} another language control (LC) model, which guides the model in generating translation in the correct language according to the instruction.
Our experiments demonstrate that this novel language control approach mitigates language confusion errors and yields a performance boost by up to $4.66$ BLEU and $8.87$ COMET scores on the MuST-C \cite{wang_covost_2021} and $4.92$ BLEU and $11.83$ COMET scores on the CoVoST-2 $\textrm{En} \rightarrow \textrm{X}$ \cite{gangi_must-c_2019}.


 
\begin{table*}[t!]
    \centering
    \caption{Examples of language confusion errors (LCE). The text in red indicates the wrong language token, which leads to the translation in the wrong language. The blue text represents the merged model that generates the correct language token.}
    \label{tab:language_confusion}
    \scalebox{0.85}{
    \begin{tabular}{llclc}
    \toprule
    \textbf{System} & \textbf{Lang.} & \textbf{LC} & \textbf{Translation example} & \textbf{LCE} ($\downarrow$) \\
    \midrule
    Reference & En $\rightarrow$ De & N/A & Und das ist in der Tat eine konservative Schätzung. & 0.00\% \\
    Task vector addition & En $\rightarrow$ De & \xmark & \textcolor{light_red}{French}: C 'est en fait une estimation conservative. & 17.97\% \\
    Task vector addition w/ Lang. Con. & En $\rightarrow$ De & \cmark & \textcolor{light_blue}{German}: Und das ist tatsächlich eine konservative Schätzung. & 0.81\% \\
    \midrule
    Reference & En $\rightarrow$ Fr & N/A & L 'évolution ne favorise pas nécessairement une vie plus longue. & 0.00\% \\
    Task vector addition & En $\rightarrow$ Fr & \xmark & \textcolor{light_red}{German}: Die Evolution schätzt nicht zwingend diejenigen, die am längsten leben. & 13.94\% \\
    Task vector addition w/ Lang. Con. & En $\rightarrow$ Fr & \cmark & \textcolor{light_blue}{French}: L 'évolution ne favorise pas nécessairement la plus longue vie.& 9.66\% \\
    \bottomrule
    \end{tabular}
    }
    \vspace{-10pt}
\end{table*}

In addition, we explore ST task vector synthesis to obtain a ST system where neither paired ST training data nor a pre-trained ST model is available using task analogies.
Similar to word analogies \cite{mikolov_efficient_2013}, the new task vector can be synthesized via task vector analogies using relevant task vectors.
Specifically, a En $\rightarrow$ Fr ST system can be synthesized using task analogies between existing machine translation (MT) in En $\rightarrow$ \{Fr, De\} and ST in En $\rightarrow$ De.
We then further demonstrate that it is possible to achieve a target language expansion on a language pair where there is no existing ST resources for this language pair.
Specifically, in this work, we aim to achieve an En $\rightarrow$ \{Fr, De\} system by merging the synthesized En $\rightarrow$ Fr ST system into an existing En $\rightarrow$ De ST system.


\section{Method}
\label{sec:method}


\begin{table}[t!]
    \centering
    \caption{Examples of generation output templates. The \textbf{bold} text indicates a language token. The \underline{underline} text represents the augmented language token, which is randomly drawn from the languages we want to merge.}
    \label{tab:instructions}
    \begin{scalebox}{0.9}{
    \begin{tabular}{lp{0.4\textwidth}}
    \toprule
    \textbf{Task} & \textbf{Output} \\
    \midrule
    MT & [SpeechGPT]: English to \textbf{German} MT \newline \textbf{German}: Ich erinnere mich an meinen ersten Feuer.\\
    \midrule
    ST  & [SpeechGPT]: English to \textbf{French} ST \newline English: i remember my first fire \newline \textbf{French}: Je me souviens de mon premier incendie.\\
    \midrule
    LC & [SpeechGPT]: English to \underline{French} ST \newline English: i remember my first fire \newline \underline{French}:\\
    \bottomrule
    \end{tabular}
    }
    \end{scalebox}
    \vspace{-10pt}
\end{table}

\subsection{Task vectors}
\label{ssec:task_vector}

First, we denote the task vector for the $i\text{-th}$ fine-tuned task as $\tau_i \in \mathbb{R}^{d}$, where $d$ is the dimension size of all model parameters.
The task vector is obtained by the difference between the $i$-th fine-tuned model parameters $\theta^{\rm FT}_i \in \mathbb{R}^{d}$ and the pre-trained model parameters $\theta^{\rm PT} \in \mathbb{R}^{d}$, i.e., $\tau_i = \theta^{\rm FT}_i - \theta^{\rm PT}$, as defined in \cite{ilharco_editing_2023}.
As shown in \cref{eq:task_vector}, the merged model parameters $\theta \in \mathbb{R}^{d}$ can be obtained by an element-wise linear interpolation between $\theta^{\rm PT}$ and $N$ different task vectors. Hyper-parameter $\lambda_i$ are scaling coefficients.

\begin{equation}\label{eq:task_vector}
    \theta = \theta^{\text{PT}} + \sum_{i=1}^{N} \lambda_i \ \tau_i.
\end{equation}
Besides this general form, we specifically focus on each weight matrix within model parameters that are typically fine-tuned in low-rank adaptation (LoRA)~\cite{hu_lora_2022} while fixing the other parameters.
Then, the model merging equation for the weight matrix corresponding to \cref{eq:task_vector} is represented as~\cite{huang_lorahub_2024}:
\begin{equation}\label{eq:task_vector_tmp}
    \mathbf{W} = \mathbf{W}^{\text{PT}} + \sum_{i=1}^{N} \lambda_i \ (\mathbf{W}_i^{\text{FT}} - \mathbf{W}^{\text{PT}}).
\end{equation}
In LoRA, each weight matrix in the self-attention module is represented as $\mathbf{W}^{\rm FT} = \mathbf{W}^{\rm PT} + \Delta \mathbf{W}$. $\Delta \mathbf{W}$ is decomposed as $\Delta \mathbf{W} = \mathbf{BA}$, where $\mathbf{B} \in \mathbb{R}^{ d_{\text{ff}} \times r}$, $\mathbf{A} \in \mathbb{R}^{r \times k}$, $d_{\text{ff}}$ is the dimension of the feedforward network, $k$ is the feature dimension, $r$ is rank, and the rank $r \ll $ min($d_{\text{ff}}$, $k$).
Therefore, by substituting the above LoRA decomposition equation into \cref{eq:task_vector_tmp}, we can obtain the following equation\footnote{Our approach is different from HuggingFace's implementation. They merge LoRA matrics in the following way $\sum_i^{N}\mathbf{B}_i \sum_i^{N}\mathbf{A}_i$.}:
\begin{equation}\label{eq:task_vector_lora}
    \mathbf{W} = \mathbf{W}^{\text{PT}} + \sum_{i=1}^N 
     \lambda_{i} \ \Delta \mathbf{W}_i
     = \mathbf{W}^{\text{PT}} + \sum_{i=1}^N 
     \lambda_{i} \ \mathbf{B}_i\mathbf{A}_i.
\end{equation}
This operation is performed for all LoRA adaptation parameters and used in our model merging stage.

\subsection{Target Language expansion via task vector addition}
\label{ssec:merging_ST}

Considering there are multiple one-to-many ST models that translate one source language $L_{s}$ into $N$ different target languages $L_{t_i} (1 \leq i \leq N)$.
We can regard one-to-one ST for each language pair as a task $\tau_i$ in \cref{eq:task_vector} and merge these ST models by addition for $i=1$ to $N$ with its task vector $\tau_{L_{s} \rightarrow L_{t_i}}^{\text{ST}}$, as follows:
\begin{equation}\label{eq:task_vector_st}
    \theta = \theta^{\text{PT}} + \sum_{i=1}^N
     \lambda_{i} \ \tau^{\text{ST}}_{L_{s} \rightarrow L_{t_i}}.
\end{equation}
The merged model would have one-to-many translation ability to translate from the source language $L_s$ to multiple target languages $L_{t_i}$.
We can further apply advanced methods of task arithmetic such as TIES-Merging \cite{yadav_ties-merging_2023}, which improves the performance by pruning task vectors and voting on their signs.

\subsection{Language Control}
\label{ssec:language_confusion}

We aim to expand language pairs in ST via simple task vector addition, as demonstrated in \cref{ssec:merging_ST}.
However, it introduces language confusion errors, where the merged model gets confused about which target language to translate into.
As we observed in a preliminary experiment, shown in \cref{tab:language_confusion}, the merged model often generates translation in the incorrect language.
In the case of being instructed to translate German, $17.97\%$ of sentences in the test split are mistranslated into French.
We hypothesize that this is caused by the lack of the capability to predict an output language according to instructions.
To address this, we propose to merge another LC model to equip language control capability, as follows,
\begin{equation}\label{eq:task_vector_st_lc}
    \theta = \theta^{\text{PT}} + \sum_{i=1}^N
     \lambda_{i} \ \tau^{\text{ST}}_{L_{s} \rightarrow L_{t_i}} + \lambda_{\text{LC}} \  \tau^{\text{LC}},
\end{equation}
where $\tau^{\text{LC}} \in \mathbb{R}^d$ is the task vector for LC and $\lambda_{\text{LC}}$ is the coefficient for LC.
The LC model can be obtained in the same instruction tuning method as ST models with different output templates (See details in \cref{ssec:instructions}).

\cref{tab:instructions} summarizes the output template when we prepare a fine-tuned model for ST and LC.
Note that the response of an ST model consists of the ASR transcript (e.g., startging from English) followed by translation text, which can be seen as chain-of-thought prompting \cite{wei_chain--thought_2022}.
As previous work has reported \cite{rubenstein_audiopalm_2023}, this helps to achieve an improved translation quality than directly generating translation.
The translation part starts with a language token (e.g., French), which corresponds to the language specified in the instruction, followed by the translation text.
The LC model is trained to generate a target language token (e.g., French) after the ASR transcript, which simulates a language token prediction part of ST given the instruction.
In the case of one-to-many ST, which we explore in this work, only source language ASR training data and ST instruction are required to train an LC model, regardless of the extended target language.
By further merging the LC model, shown in \cref{fig:merging}, we expect the merged model to have enhanced the capability to control a target language based on the instruction.

\subsection{ST task vector synthesis via task analogies}
\label{ssec:language_vector}

Consider the scenario where one wants to expand a language pair in an existing ST model, but no paired ST data nor pre-trained ST model is available for the pair.
For this purpose, we propose synthesizing an ST task via task analogies formulated in task arithmetic \cite{ilharco_editing_2023}.
For tasks having the relationship $\text{task}_1 : \text{task}_2 :: \text{task}_3 : \text{task}_4$, these task vectors holds the following equation:
\begin{equation}\label{eq:task_analogies}
 \tau_{\text{task}_4} = \tau_{\text{task}_3} +\tau_{\text{task}_2} - \tau_{\text{task}_1}.
\end{equation}
Inspired by pivot translation \cite{kauers_interlingua_2002,allman_catalan-english_2006}, we introduce a pivot language $L_p$ where a pre-trained ST model is available.
Using task analogies, task analogies between MT of $L_s \rightarrow L_p$ (\textbf{available}), MT of $L_s \rightarrow L_t$  (\textbf{available}), ST of $L_s \rightarrow L_p$  (\textbf{available}) and ST of $L_s \rightarrow L_t$ (\textbf{not available}) are written as:
\begin{equation}\label{eq:our_analogies}
\tau_{L_s \rightarrow L_p}^{\text{MT}} : \tau_{L_s \rightarrow L_t}^{\text{MT}} :: \tau_{L_s \rightarrow L_p}^{\text{ST}} : \tau_{L_s \rightarrow L_t}^{\text{ST}}.
\end{equation}
Following \cref{eq:task_analogies}, we can synthesize ST of $L_s \rightarrow L_t$ via task analogies as,
\begin{equation}\label{eq:task_analogies_lang_vector}
    \tau_{L_s \rightarrow L_t}^{\text{ST}} =
    \lambda_{ST} \ \tau_{L_s \rightarrow L_p}^{\text{ST}} +
    \lambda_{MT} \ (\tau_{L_s \rightarrow L_t}^{\text{MT}} - \tau_{L_s \rightarrow L_p}^{\text{MT}}),
\end{equation}
where $\lambda_{ST}$ and $\lambda_{MT}$ are the merging coefficients.
The merged model would have the ability to translate from the source language $L_s$ to target languages $L_{t}$ that we want to expand.
Finally, we further merge this synthesized ST model into the existing ST model for language expansion, as in \cref{ssec:merging_ST}.
\begin{figure}[t!]
    \centering
    \includegraphics[width=0.4\textwidth]{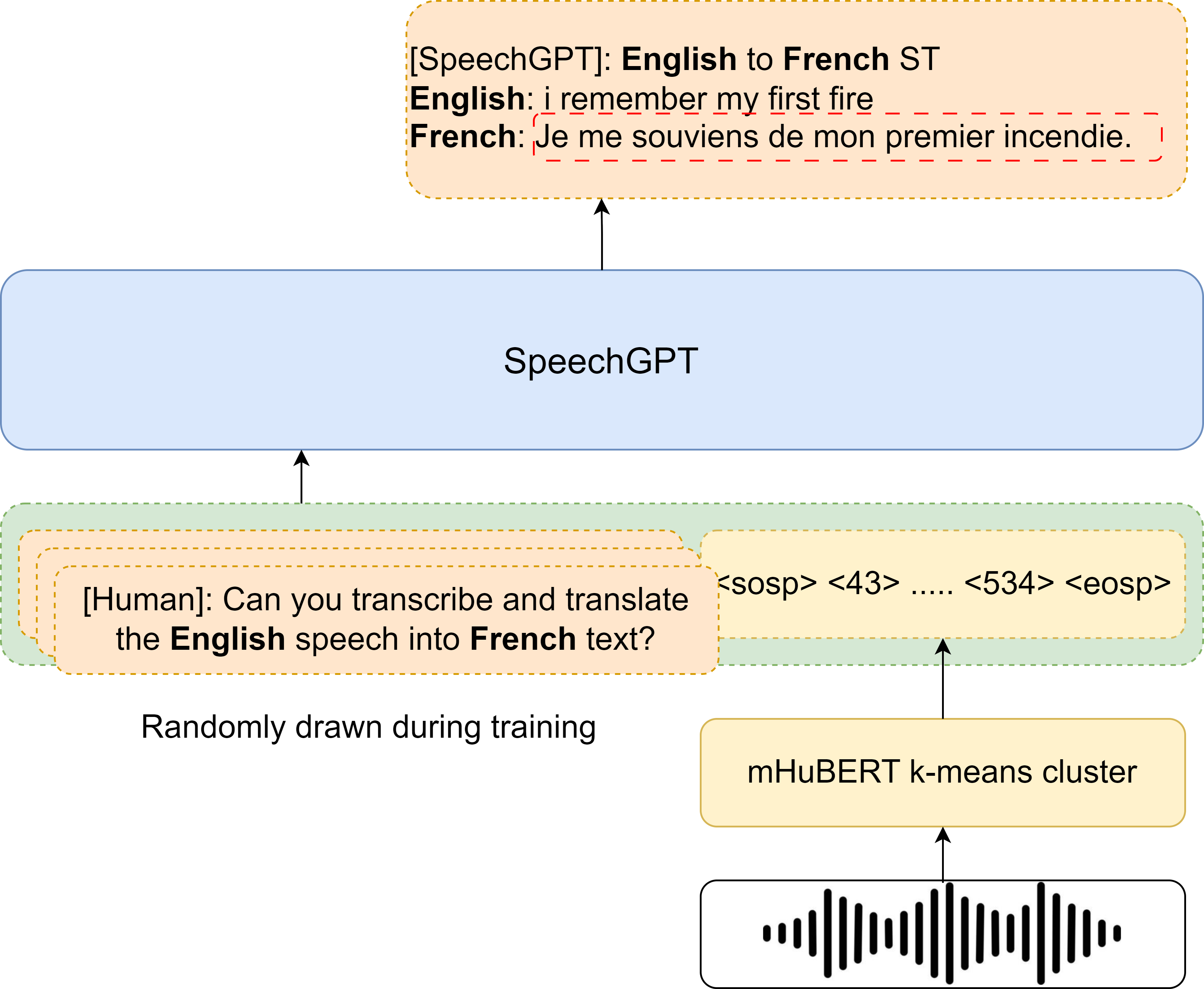}
    \caption{The overview of fine-tuning process. The instruction is randomly drawn during training. Speech discrete units are obtained via the mHuBERT k-means cluster.\protect\footnotemark The input text after the special token [Human] indicates the natural language instruction. The output text (transcript and translation) starts with a special token [SpeechGPT].}
    \label{fig:system}
    \vspace{-15pt}
\end{figure}

\footnotetext{\url{https://dl.fbaipublicfiles.com/hubert/mhubert_base_vp_en_es_fr_it3.pt}}

\section{Experiments}
\label{sec:experiments}

\subsection{Fine-tuning}
\label{ssec:fine_tuning}

\cref{fig:system} summarizes the detailed fine-tuning flow.
We used SpeechGPT-7B-cm\footnote{\url{https://huggingface.co/fnlp/SpeechGPT-7B-cm}}, a large speech-text multi-modal pre-trained model with 7B parameters, as our backbone \cite{zhang_speechgpt_2023} for ST, MT, and LC tasks.
We prepared pre-trained ST, MT, and LC models by SpeechGPT’s instruction tuning with LoRA \cite{hu_lora_2022} as discussed in \cref{ssec:task_vector} by updating key, query, value, and output projections with the rank $r=32$.
We used $\textrm{En} \rightarrow \textrm{De}$ and $\textrm{En} \rightarrow \textrm{Fr}$ of MuST-C \cite{gangi_must-c_2019}, and also $\textrm{En} \rightarrow \textrm{De}$ and $\textrm{En} \rightarrow \textrm{Zh}$ of CoVoST-2 \cite{wang_covost_2021}.
For fine-tuning, we used $4$ A6000 GPUs with batch size $128$, learning rate $2\times10^{-5}$.
We used AdamW \cite{loshchilov_decoupled_2019} to update $4200$ steps for MuST-C \cite{gangi_must-c_2019} and $2100$ steps for CoVoST-2 \cite{wang_covost_2021} experiments.
We set the merging coefficients $\lambda$ between $0.2$ and $1.3$ for MT/ST/LC models.
All the hyper-parameters were searched according to development set performance.

\subsection{Instructions}
\label{ssec:instructions}

The input instructions for ST are constructed as follows: `Can you transcribe and translate \ldots,’ and randomly selected from $10$ templates during fine-tuning, as done in SpeechGPT \cite{zhang_speechgpt_2023}.
For the output template, we follow AudioPaLM’s style \cite{rubenstein_audiopalm_2023} to instruct the model to generate a transcript and then translation.
For MT used in task analogies experiments, we changed the input instructions to `Can you translate \ldots.’ and only instructed to generate translation.
For the LC task (see \cref{ssec:language_confusion}), we use ASR training data from $\textrm{En} \rightarrow \textrm{Fr}$ for MuST-C experiments and $\textrm{En} \rightarrow \textrm{De}$ for CoVoST-2 experiments.
Specifically, the input instructions for LC are adapted from ST with an augmented target language token (\underline{underline} text in \cref{tab:instructions}), where the target language is randomly drawn from the languages we want to merge.

\section{Results}
\label{sec:results}

\subsection{Target Language expansion}
\label{ssec:language_expansion}

Target language expansion merges one-to-one ST models in different language pairs into a single one-to-many ST model.
This work focuses on merging two $\textrm{En} \rightarrow \textrm{X}$ ST models.
We compare our merged models with topline approaches: joint fine-tuning, which jointly trains the ST model on the combined datasets of two language pairs, and monolingual fine-tuning, which fine-tunes the ST model on each language pair.
We show that our proposed method can be incorporated with the advanced task arithmetic method, TIES-Merging \cite{yadav_ties-merging_2023}, which aims to reduce parameter interference by pruning the parameters and aggregating task vectors.
For evaluation, we use BLEU score \cite{papineni_bleu_2001} as the primary and COMET\footnote{\url{https://huggingface.co/Unbabel/wmt22-comet-da}} \cite{rei_comet_2020} as the auxiliary metric.

\begin{table}[t!]
    \centering
    \caption{Performance of target language expansion experiment on Must-C. Joint fine-tuning (topline) requires re-training of both datasets, which we want to avoid because of costs. We use \underline{underline} and \textbf{bold} to denote the best performers in merged models for BLEU and COMET scores.}
    \label{tab:must_c_2_lang}
    \scalebox{0.75}{
    \begin{tabular}{lccccc}
        \toprule
        \multirow{2}{*}{\textbf{Name}} &
            \multicolumn{2}{c}{\textbf{En $\rightarrow$ De}} &
            \multicolumn{2}{c}{\textbf{En $\rightarrow$ Fr}} \\
        \cmidrule{2-3}\cmidrule{4-5}
        & BLEU & COMET & BLEU & COMET \\  
        \midrule
        (A0) SpeechGPT & 0.00 & 36.28 & 0.00 & 36.28 \\
        (A0$^\prime$) \quad + LC & 0.00 & 36.28 & 0.00 & 36.28 \\
        \midrule
        (A1) Joint fine-tuning (topline) & 18.40 & 71.51 & 27.16 & 72.51 \\
        (A1$^\prime$) \quad + LC & 18.33 & 71.39 & 26.45 & 70.84 \\
        \midrule
        (A2) En $\rightarrow$ De & 19.94 & 72.44 & N/A & N/A \\
        (A2$^\prime$) \quad + LC & 19.78 & 72.28 & N/A & N/A \\
        \midrule
        (A3) En $\rightarrow$ Fr & N/A & N/A & 27.64 & 73.11 \\
        (A3$^\prime$) \quad + LC & N/A & N/A & 27.87 & 72.81 \\
        \midrule
        (A4) Task vector addition \scriptsize{(A2) + (A3)} & 11.75 & 61.22 & 17.46 & 63.64 \\
        (A5) \quad + LC & 15.52 & 67.09 & 17.73 & 66.73 \\
        \midrule
        (A6) TIES$^\ddagger$ \scriptsize{(A2) + (A3)} & 0.36 & 40.23 & 0.00 & 39.18 \\
        (A7) \quad + LC & \underline{16.41} & \textbf{70.09} & \underline{20.22} & \textbf{69.44} \\
        \bottomrule
        \end{tabular}
    }
    \vspace{-10pt}
\end{table}

\begin{table}[t!]
    \centering
    \caption{Performance of target language expansion experiment on CoVoST-2.}
    \label{tab:covost_2langs}
    \scalebox{0.75}{
        \begin{tabular}{lcccc}
        \toprule
        \multirow{2}{*}{\textbf{Name}} &
            \multicolumn{2}{c}{\textbf{En $\rightarrow$ De}} &
            \multicolumn{2}{c}{\textbf{En $\rightarrow$ Zh$^*$}} \\
        \cmidrule{2-3}\cmidrule{4-5}
        & BLEU & COMET & BLEU & COMET \\ 
        \midrule
        (B1) Joint fine-tuning (topline) & 17.06 & 67.68 & 15.96 & 91.13 \\
        \midrule
        (B2) En $\rightarrow$ De & 18.52 & 69.11 & N/A & N/A \\
        \midrule
        (B3) En $\rightarrow$ Zh & N/A & N/A & 18.03 & 92.04 \\
        \midrule
        (B4) Task vector addition \scriptsize{(B2) + (B3)} & 13.60 & 64.91 & 9.56 & 78.32\\
        (B4$^\prime$) \quad + LC & 13.48 & \textbf{67.28} & 11.57 & 86.18 \\
        \midrule
        (B6) TIES$^\ddagger$ \cite{yadav_ties-merging_2023} \scriptsize{(B2) + (B3)} & 8.27 & 54.05 & 13.19 & 83.85 \\
        (B6$^\prime$) \quad + LC & \underline{13.95} & 66.18 & \underline{14.48} & \textbf{90.15} \\
        \bottomrule
        \end{tabular}
    }
    \vspace{-15pt}
\end{table}

\cref{tab:must_c_2_lang} shows the results of the target language expansion experiment on MuST-C between $\textrm{En} \rightarrow \textrm{De}$ and $\textrm{En} \rightarrow \textrm{Fr}$ ST models.
Our experiments on MuST-C, shown in \cref{tab:must_c_2_lang}, show that task vector addition (A4) achieves BLEU scores of $11.75$ (\textrm{En} $\rightarrow$ \textrm{De}) and $17.46$ (\textrm{En} $\rightarrow$ \textrm{Fr}), but suffers from language confusion errors.
Our proposed LC model could eliminate such errors, from $17.97\%$ to $0.81\%$, $13.94\%$ to $9.66\%$, leading to better BLEU scores: $15.52$ and $17.73$ (A5).
Similarly, we observe the same trend of performance improvement in COMET scores.
In addition, TIES-Merging led to severe language confusion (A6).
We hypothesize that this is due to the pruning step in TIES-Merging, where the merged model loses its instruction-following capability
However, the LC model can equip the forgotten capability, resulting in the best performance: $16.41$ and $20.22$ in BLEU, $70.09$, and $69.44$ in COMET scores (A7).
Compared to the joint fine-tuning approach (A1), our best-merged model (A7) has slight performance degradation, yet requires no training on combined datasets and supports multiple language pairs in a single model.
Additionally, it translates into two language pairs within one model rather than two separated monolingual fine-tuned models ((A2) and (A3)).

To examine whether the performance gains stem purely from eliminating language confusion errors by merging the LC model, we performed an ablation study, presented in \cref{tab:must_c_2_lang}.
As we can see, vanilla SpeechGPT (A0) does not possess inherent translation capability on MuST-C.
Adding the LC model does not improve vanilla SpeechGPT (A0$^\prime$ vs A0) because no language confusion errors occur in vanilla SpeechGPT (A0).
Similarly, adding the LC model to fine-tuned baselines that do not exhibit language confusion errors does not yield performance gains (A1$^\prime$ vs. A1, A2$^\prime$ vs A2, and A3$^\prime$ vs A3).
On the other hand, the results demonstrate that the LC model consistently enhances performance for merged models (A5 vs. A4 and A7 vs. A6), where language confusion errors occur.
This empirical evidence suggests that the performance improvements achieved with our proposed LC model stem from eliminating language confusion errors rather than other factors.

We also show the results of the experiments on different corpus CoVoST-2, for different languages $\textrm{En} \rightarrow \textrm{De}$ and $\textrm{En} \rightarrow \textrm{Zh}$ in \cref{tab:covost_2langs}.
Although we observed fewer language confusion errors in this setting than in merging two European languages on MuST-C, we still consistently improved by using the proposed LC model.

\subsection{ST task vector synthesis}
\label{ssec:language_adaptation}

In this section, we explored the possibility of expanding a language pair from an existing ST model in a scenario where neither paired ST data nor a pre-trained model was available, as described in \cref{ssec:language_vector}.
Specifically, we first synthesized the ST model in the target language pair using task analogies (discussed in \cref{ssec:language_adaptation}).
Next, we merged this synthesized ST model with the existing ST model via task vector addition.
We aim to synthesize a new ST pair of En $\rightarrow$ Fr (\textbf{not available}) by leveraging MT of En $\rightarrow$ De (\textbf{available}), MT of $\rightarrow$ Fr (\textbf{available}), and ST of En $\rightarrow$ De (\textbf{available}).
Available models were trained on the MuST-C dataset.

We summarize the results of the target language expansion with synthesized En $\rightarrow$ Fr ST in \cref{tab:synthesized_st}.
We find the direct application of task analogy failed due to severe language confusion errors (C4) in \cref{tab:synthesized_st}.
By further merging our proposed LC model, the synthesized model (C5) yielded BLEU and COMET scores of $21.65$ and $69.62$ in $\textrm{En} \rightarrow \textrm{Fr}$, following the trend in target language expansion experiments in \cref{ssec:language_expansion}.
Notably, we can obtain the ST model without relying on paired ST data with competitive performance, which only shows $21.67\%$ and $4.77\%$ relative performance degradation in BLEU and COMET scores compared to the model fine-tuned on ST data ((A3) in \cref{tab:must_c_2_lang}).
Once we obtain the synthesized ST model for a language we aim to expand, we merge it into the existing ST model for target language expansion via TIES-Merging ((C6) and (C7)).
Despite it only achieved BLEU scores of $9.22$ and $7.16$ in $\textrm{En} \rightarrow \textrm{De}$ and $\textrm{En} \rightarrow \textrm{Fr}$, this shows the potential of target language expansion without using the ST data.
We believe that performance degradation was introduced by merging models multiple times, which we leave for future work.

\section{Conclusion}
\label{sec:conclusion}

\begin{table}[t!]
    \centering
    \caption{Performance of target language expansion with the synthesized ST for $\textrm{En} \rightarrow \textrm{Fr}$.}
    \label{tab:synthesized_st}
    \scalebox{0.75}{
    \begin{tabular}{lccccc}
        \toprule
        \multirow{2}{*}{\textbf{Name}} &
            \multicolumn{2}{c}{\textbf{En $\rightarrow$ De}} &
            \multicolumn{2}{c}{\textbf{En $\rightarrow$ Fr}} \\
        \cmidrule{2-3}\cmidrule{4-5}
        & BLEU & COMET & BLEU & COMET \\  
        \midrule
        \multicolumn{4}{c}{\textbf{MT}} \\
        \midrule
        (C1) En $\rightarrow$ De & 24.82 & 80.95 & N/A & N/A \\
        (C2) En $\rightarrow$ Fr & N/A & N/A & 35.85 & 82.23 \\
        \midrule
        \multicolumn{4}{c}{\textbf{ST}} \\
        \midrule
        (C3) En $\rightarrow$ De & 19.94 & 72.44 & N/A & N/A \\
        (C4) \multirowcell{2}[0pt][l]{Task analogy \scriptsize{(C3) + (C2) - (C1)} \\(Synthesized En $\rightarrow$ Fr)} & \multirow{2}{*}{N/A} & \multirow{2}{*}{N/A} & \multirow{2}{*}{0.22} & \multirow{2}{*}{31.30} \\\\
        \midrule
        \multirow{2}{*}{(C5)} \multirowcell{2}[0pt][l]{(C4) + LC \\(Synthesized En $\rightarrow$ Fr)} & \multirow{2}{*}{N/A} & \multirow{2}{*}{N/A} & \multirow{2}{*}{\underline{21.65}} & \multirow{2}{*}{\textbf{69.62}} \\\\
        \midrule
        (C6) TIES$^\ddagger$ \cite{yadav_ties-merging_2023} \scriptsize{(C3) + (C5)} & 0.00 & 25.94 & 0.00 & 27.30 \\
        (C7) \quad + LC & 9.22 & 66.84 & 7.16 & 61.10 \\
        \bottomrule
        \end{tabular}
    }
    \vspace{-15pt}
\end{table}
\blfootnote{$^*$ We report character-level BLEU scores on Chinese translation.}
\blfootnote{$^\ddagger$ We prune $50\%$ of parameters in TIES-Merging experiments.}
This paper aims to expand the target language support in the existing one-to-one ST model via task arithmetic.
We find that the direct application causes language confusion, generating translations in incorrect languages.
Our proposed language language control model effectively eliminates this issue, enabling accurate target language generation.
In addition, we demonstrate that task analogies can achieve ST task synthesis even when without paired speech translation data.

\clearpage

\printbibliography[heading=bibnumbered]

\end{document}